# Arabic Little STT: Arabic Children Speech Recognition Dataset

Mouhand Alkadri, Dania Desouki, Khloud Al Jallad

Department of Information and Communication Engineering,

Arab International University, Daraa, Syria.

* Corresponding author. E-mail(s): k-aljallad@aiu.edu.sy, 201820115@aiu.edu.sy, 202020003@aiu.edu.sy

## Abstract

The performance of Artificial Intelligence (AI) systems fundamentally depends on high-quality training data. However, low-resource languages like Arabic suffer from severe data scarcity. Moreover, the absence of child-specific speech corpora is an essential gap that poses significant challenges. To address this gap, we present our created dataset, Arabic Little STT, a dataset of Levantine Arabic child speech recorded in classrooms, containing 355 utterances from 288 children (ages 6–13). We further conduct a systematic assessment of Whisper, a state-of-the-art automatic speech recognition (ASR) model, on this dataset and compare its performance with adult Arabic benchmarks. Our evaluation across eight Whisper variants reveals that even the best-performing model (Large-v3) struggles significantly, achieving a 66% word error rate (WER) on child speech – starkly contrasting with its sub-20% WER on adult datasets. These results align with other research on English speech. Results highlight the critical need for dedicated child speech benchmarks and inclusive training data in ASR development. Emphasizing that such data must be governed by strict ethical and privacy frameworks to protect sensitive child information We hope that this study provides an initial step for future work on equitable speech technologies for Arabic-speaking children, enriching the children's demographic representation in ASR datasets.

# 1. Introduction

Automatic Speech Recognition (ASR) has been a hot research topic in NLP in the last several years. Beyond its importance in research, its capabilities extend from powering virtual assistants, voice-controlled devices to enabling hands-free data entry and improving accessibility for kids and for adults with disabilities. Notably, ASR is also playing an increasingly transformative role in the online education landscape. ASR is heavily used in content transcription, which does not only result in a more streamlined education process, but also creates equal opportunities for content comprehension for students with hearing disabilities. Furthermore, ASR systems offer significant advantages for language learning applications, especially those that evaluate pronunciation and provide feedback. These important applications of ASR in EdTech highlight a profound opportunity to enhance the online educational process for children, who naturally tend to interact with the world through speech rather than text. This preference for speech-based interaction highlights the critical need for accurate ASR tailored to children for an effective learning experience.

ASR task has large-scale datasets, such as TED-LIUM (Hernandez et al., 2018; Rousseau et al., 2012, 2014), LibriSpeech (*Librispeech: An ASR Corpus Based on Public Domain Audio Books | IEEE Conference Publication | IEEE Xplore*, n.d.) and CommonVoice (Ardila et al., 2020), as well as sophisticated models, such as FastConformer-Hybrid PCD (*[2507.13977] Open Automatic Speech Recognition Models for Classical and Modern Standard Arabic*, n.d.), Whisper (Radford et al., n.d.), and Wav2Vec (Schneider et al., 2019).

Although ASR has large-scale datasets and sophisticated models, most of them target adult speech. Thus, most publicly available datasets lack children's voices. This gap increases in low-resources languages such as Arabic, where dialectal diversity, morphological complexity, and the absence of child-specific speech corpora is an essential gap that poses significant challenges. To overcome this gap, we created a dataset for children's voices and named it Arabic Litte STT[1]. Moreover, we evaluated the Whisper (Radford et al., n.d.)

---

[1] Dataset available at this link https://huggingface.co/datasets/little-stt/little-stt-dataset

models on our newly developed dataset of Arabic-speaking children's voices, comparing its performance with performance on adult speech datasets.

## 2. Related Works

### 2.1. Speech Datasets

Automatic Speech Recognition (ASR) models rely heavily on high-quality datasets. For widely represented languages like English, numerous datasets exist, such as LibriSpeech (*Librispeech: An ASR Corpus Based on Public Domain Audio Books | IEEE Conference Publication | IEEE Xplore*, n.d.) and the three versions of TED-LIUM (Hernandez et al., 2018; Rousseau et al., 2012, 2014).

There are many multilingual datasets, such as Common Voice (Ardila et al., 2020) and FLEURS (Conneau et al., 2022) include underrepresented languages like Arabic.

Moreover, there are dedicated Arabic datasets like Casablanca dataset (Talafha et al., 2024) which represents different Arabic dialects.

Table 1 shows a comparison between some available ASR datasets.

| Dataset | Size | Audio Source | Arabic Representation |
|---|---|---|---|
| **LibriSpeech** | 1K hours | Read speech (Audiobooks) | No |
| **TED-LIUM (Release 1)** | 118 hours | TED Talks | No |
| **TED-LIUM (Release 2)** | 207 hours | TED Talks | No |
| **TED-LIUM (Release 3)** | 452 hours | TED Talks | No |
| **Common Voice V17[2]** | 31175 hours total, 20408 validated hours | Crowdsourced read speech (various sources like blogs, books, movies) | 157 Hours |
| **FLEURS** | ~12 hours per language | Read speech (based on n- | ~12 hours |

---
[2] https://huggingface.co/datasets/mozilla-foundation/common_voice_17_0

| | (102 language) | way parallel sentences from FLoRes-101 MT benchmark) | |
| --- | --- | --- | --- |
| **Casablanca** | ~48 hours | YouTube episodes from TV series | ~48 hours (across 8 dialects) |

*Table 1. Comparison between some available ASR datasets*

However, none of the previously mentioned datasets take children's speech into consideration, a gap that disproportionately affects underrepresented languages and leaves ASR systems ill-equipped to handle the acoustic and linguistic nuances of younger speakers.

2.2. Children's Speech Dataset

There are few available resources for children's speech, such as the CMU Kids Corpus (*The CMU Kids Corpus - Linguistic Data Consortium*, n.d.) 5,180 utterances from 76 English-speaking children, PF_STAR (*(PDF) The PF_STAR Children's Speech Corpus*, n.d.) 60 hours of non-native English from European children, and MyST (Pradhan et al., 2023) 400 hours of conversational English.

Moreover, these datasets are exclusively English-centric.

Although it is a linguistically diverse language, Arabic has no dedicated resources for children's speech. Table 2 shows a comparison between some available children ASR datasets.

| **Dataset** | **Size** | **Audio Source** |
| --- | --- | --- |
| **CMU Kids Corpus** | 9 hours | Read-aloud sentences by children |
| **PF_STAR** | 60 hours | Recorded as part of the EU FP5 PF STAR project |
| **MyST** | 400 hours | Recorded sessions as part of My Science Tutor project |

*Table 2. Comparison between available children specific ASR datasets*

2.3. Child-Specific ASR Adaptations

Efforts to optimize ASR systems for children's speech have focused on English. Jain et al. adapted Whisper to English children speech (Jain et al., 2023). Shi et al. applied Test-Time adaptation (TTA) methods—A process allows adult targeted ASR models to continuously adapt to each child speaker at test time—to improve ASR system on children's voices (Shi et al., 2024). These advances highlight the potential for adaptation but remain inaccessible to low-resource languages due to data scarcity.

2.4. ASR State-of-the-Art Models

Significant advancements in Automatic Speech Recognition (ASR) have been largely driven by open-source models. Notable examples include FastConformer-Hybrid PCD (*[2507.13977] Open Automatic Speech Recognition Models for Classical and Modern Standard Arabic*, n.d.), Whisper (Radford et al., n.d.) a transformer-based multilingual model, Whisper-Medusa (Segal-Feldman et al., 2024) an optimized variant of Whisper to accelerate inference, Canary (Puvvada et al., 2024) an ASR model that Surpasses Whisper in low-resource languages and Wav2Vec (Schneider et al., 2019) a pioneer in unsupervised pre-training for speech recognition.

Table 3 shows a comparison between the state-of-the-art ASR models.

| Model | Architecture | Supports Arabic |
|---|---|---|
| **FastConformer-Hybrid PCD** | Conformer Based | Yes |
| **Whisper** | Transformer Based | Yes (Multilingual Variant) |
| **Whisper-Medusa** | Transformer Based | No |
| **Canary** | Transformer Based | No |
| **Wav2Vec** | Convolution Based | Yes (XLS-R Variant) |

*Table 3. Comparison between the state-of-the-art ASR models*

Comparing these models on the Open Universal Arabic ASR Leaderboard (Wang et al., 2024) which benchmarks open-source models for Arabic ASR tasks, reveals that Whisper large v3 achieved an average WER of 36.86. As of August 16, 2025, this placed it third on the leaderboard. NVIDIA's FastConformer models occupied the top two positions, achieving the best scores using pure greedy decoding and language model (LM) integration, respectively.

## 3. Our Created Dataset (Arabic Little STT)

Collecting child speech data for ASR poses unique challenges, as it requires navigating parental consent and institutional approvals. These factors contribute to the scarcity of Arabic child speech corpora. To address this scarcity and manage Arabic's inherent linguistic complexity, we refined our scope to focus specifically on the Levantine (Shami) dialect, a widely used dialect in the region of Levant, and more specifically we covered the Syrian variant of the Levantine dialect. Our resulting dataset comprises 355 utterances recorded from 288 children (157 male, 131 female), aged 6 to 13 years.

Aiming to reflect a realistic ASR use case, all audio was captured in classroom environments using standard smartphone microphones. This setting naturally introduced moderate levels of ambient noise, including sounds such as keyboard clicks and peer conversations. Recordings were stored in the AAC format, and later converted to WAV. The average duration of individual audio clips is approximately 10 seconds, accompanied by transcriptions averaging 13 words per utterance (observed range: 3 to 62 words).

Figure 1 is a stacked bar chart of the age and gender of the participating children, where it shows that most children were of ages 8 and 9. The minimum age group was 12 as we have only 6 children. Participating kids were more males than females in most ages.

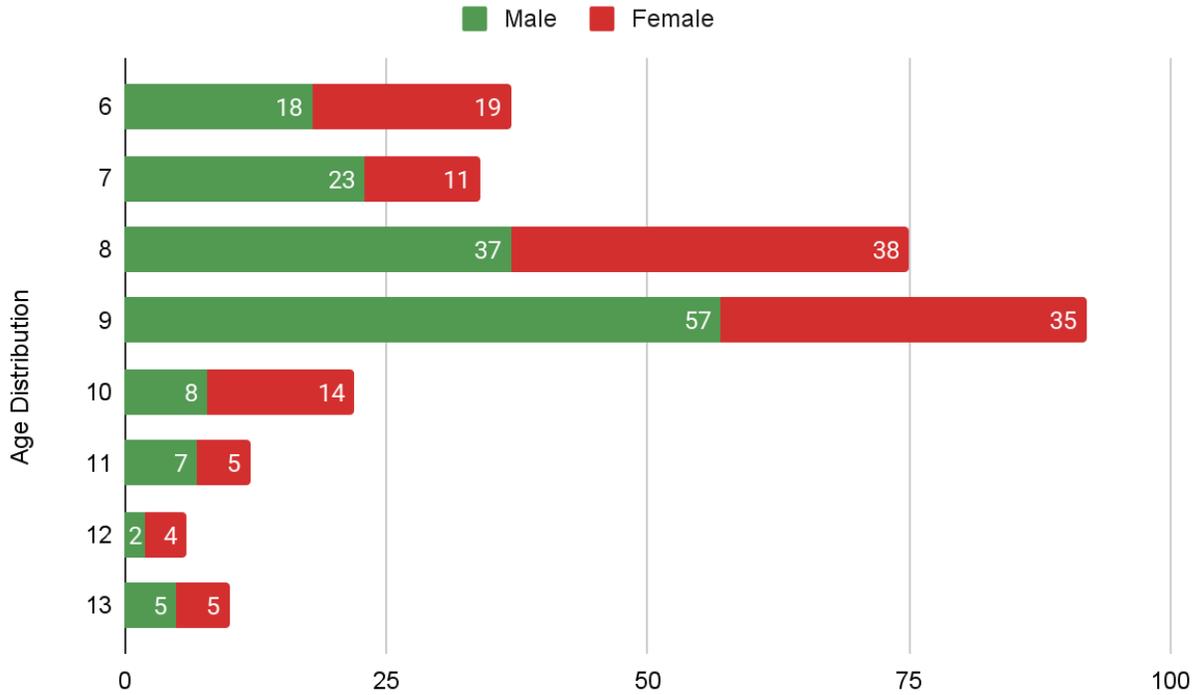

*Figure 1. A stacked bar chart of the age and gender of the participated children*

Transcription content was deliberately centered on themes of programming, robotics, and artificial intelligence, directly mirroring the Information Technology (IT) classroom context in which the recordings were conducted.

*Table 4. Samples from transcripts in our dataset.*

| |
|---|
| يعني ممكن يصير في روبوتات او اغراض الكترونية تساعدك بالأدوات المنزلية |
| بدي اخترع روبوت بنضف وبساوي اكل مشان ما نتعب حالنا |
| مثلا سيارة تيسلا بتسوق لحالها بس تحددلها الموقع بتروح لحالها بتبطىء السرعة وبتزود |
| أنو الناس يلي برمجو نسيو شغلة او حطو شغلة زيادة |
| انو البيت بالمستقبل بكون مثل الآلي بنضف البيت لبين ما روح واجي وبساوي الأكل لبين ما روح واجي |

*Table 4. Samples from transcripts in our dataset*

## 3.1. Audio Preprocessing

Following the initial collection phase, each record was manually verified for audibility (e.g., sufficient volume, minimal background noise) and discarded if it was not audible for a human. The resulting filtered set was manually transcribed following Whisper's non-English conventions (Radford et al., n.d.) All transcribers were native Levantine (Shami) Arabic speakers ensuring accuracy and language context awareness.

To further standardize the evaluation, Arabic-specific normalization was applied simplifying the Arabic writing into a commonly used form, while preserving the meaning for the human reader:

- **Diacritic Removal** Stripped Harakat (Fatha: - فتحة \َ\, Damma: - ضمة \ُ\, Kasra: - كسرة \ِ\, Sukun: - سكون \ْ\) and Shadda (- شدة \ّ\).
- **Tatweel Elimination**: Removed elongated characters (e.g., ـ) to align with modern Arabic text standards.
- **Alef Normalization** Unified آ, إ, أ into ا, reducing orthographic variability.

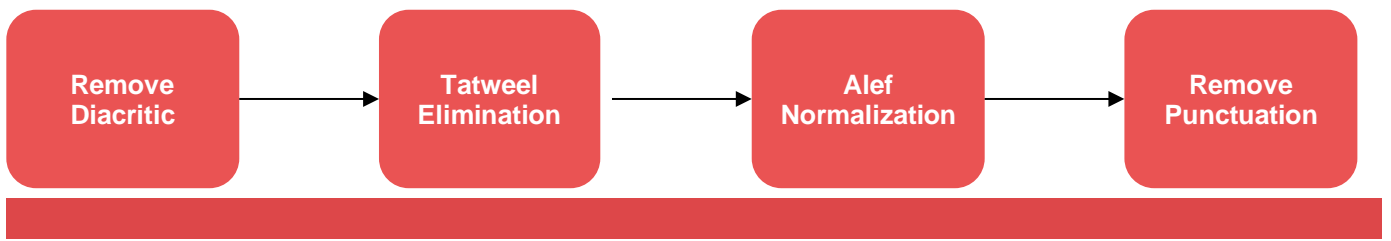

*Figure 2. An illustration of the pipeline used to normalize Arabic transcription*

## 4. Evaluation Experiment

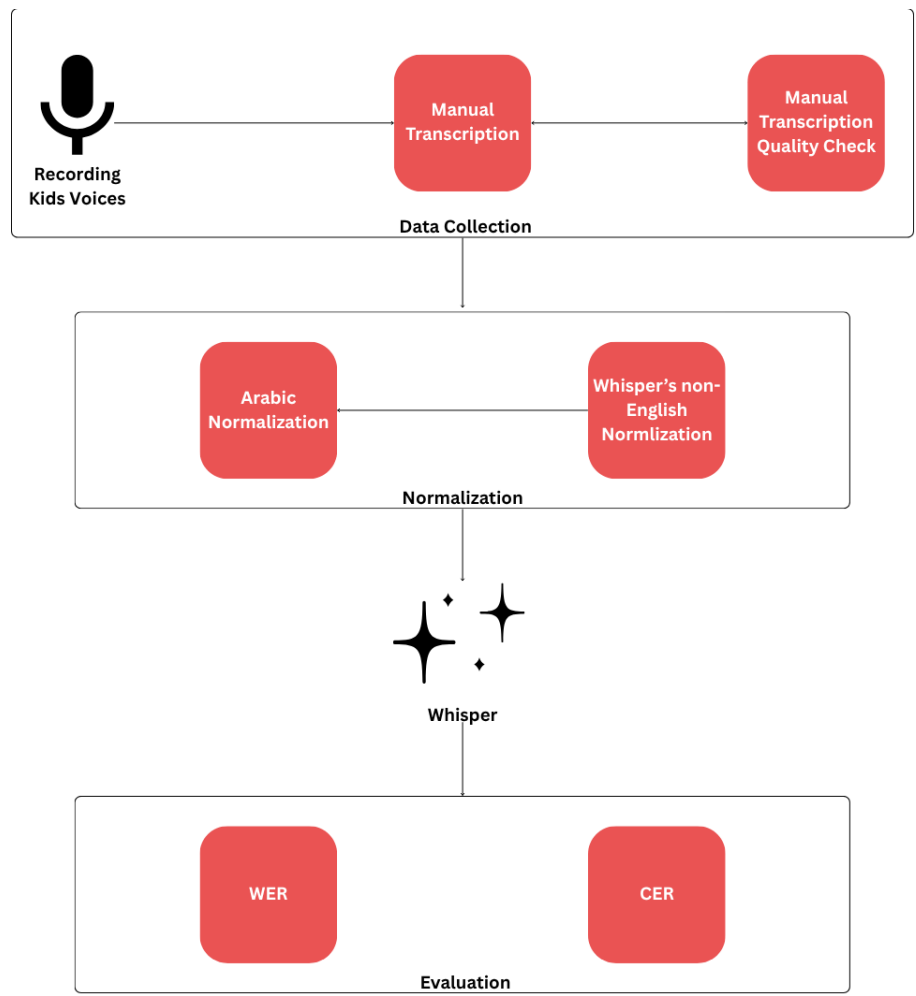

*Figure 3. A block diagram highlighting the full pipeline*

After exploring different state-of-the-art Automatic Speech Recognition (ASR) models, we chose the Whisper models family for testing our dataset. The decision was made for two reasons:

- **Proven Strong Performance in Arabic:** Whisper models perform very well with Arabic speech. As supported by its high ranking (third place) on the Open Universal Arabic ASR Leaderboard (Wang et al., 2024)

- **Availability of Similar Published Results for Children's Speech:** Researchers (Jain et al.) have already tested Whisper's performance on datasets of English-speaking children. As this evaluation exists, we wanted to use Whisper on our Arabic children's dataset. This allows us to see if the patterns and challenges found when recognizing English children's speech also appear when recognizing Arabic children's speech.

## 4.1 Whisper

Whisper (Radford et al., n.d.) is a Transformer based encoder-decoder model, also referred to as a sequence-to-sequence model. The Whisper models are trained on over 680,000 hours of audio and the corresponding transcripts collected from the internet, 65% of this data (or 438,218 hours) represents English-language audio and matched English transcripts, roughly 18% (or 125,739 hours) represents non-English audio and English transcripts, while the final 17% (or 117,113 hours) represents non-English audio and the corresponding transcript. This non-English data represents 98 different languages, Arabic is one of which with 739 hours of transcribed audio.

Multiple variants of the Whisper model were released with sizes ranging from tiny (39M parameters) to Large (1.5B parameters), some of which are released into two different settings, Multilingual and English only. The multilingual models were trained on both speech recognition and speech translation. During training, tasks are distinguished using special tokens such as <|transcribe|> for speech recognition and <|translate|> for translation into English. Table 5 shows a comparison between whisper variants

| Variant | Number of Parameters | English-only Model Available | Release Date |
|---|---|---|---|
| Tiny | 39 M | Yes (tiny.en) | September 2022 |
| Base | 74 M | Yes (base.en) | September 2022 |
| Small | 244 M | Yes (small.en) | September 2022 |
| Medium | 769 M | Yes (medium.en) | September 2022 |
| Large | 1550 M | No | September 2022 |
| Large v2 | 1550 M | No | December 2022 |
| Large v3 | 1550 M | No | November 2023 |
| Large v3 Turbo | 809 M | No | November 2023 |

*Table 5. Comparison between Whisper variants*

## 4.2. Experiment Setup

Eight Whisper model variants—Tiny[3], Base[4], Small[5], Medium[6], Large[7], Large-v2[8], Large-v3[9], and Large-v3 Turbo[10]— were evaluated spanning parameter sizes from 39M (Tiny) to 1.5B (Large-v3). This coverage reflects diverse deployment scenarios: smaller models (e.g., Tiny) suit resource-constrained edge devices, while larger variants (e.g., Large-v3) target high-accuracy cloud-based applications given that ASR systems operate in diverse environments with varying computational limits (e.g., mobile app, servers). By testing Whisper's full spectrum of sizes, we provide actionable insights for developers balancing speed, memory, and accuracy trade-offs in child-centric applications.

Regarding methodology, as Whisper models are multi-task models capable of transcription and translation (Radford et al., n.d.), the experiment was conducted using transcription mode, more specifically language detection and transcription as no language prompt was provided to the model, only relying on the model's built-in language detection. Furthermore, to ensure a fair comparison, the model's output was normalized using the same steps described earlier for our dataset's transcription.

Finally, evaluation was measured using two standard ASR metrics: Word Error Rate (WER) and Character Error Rate (CER) (Srivastav et al., 2023; Wang et al., 2024)

## 4.3. Results

All evaluated Whisper variants demonstrated strong language detection accuracy, particularly the Large variant, which achieved near-perfect detection (>99%) for Arabic (Figure 5).

---

[3] https://huggingface.co/openai/whisper-tiny
[4] https://huggingface.co/openai/whisper-base
[5] https://huggingface.co/openai/whisper-small
[6] https://huggingface.co/openai/whisper-medium
[7] https://huggingface.co/openai/whisper-large
[8] https://huggingface.co/openai/whisper-large-v2
[9] https://huggingface.co/openai/whisper-large-v3
[10] https://huggingface.co/openai/whisper-large-v3-turbo

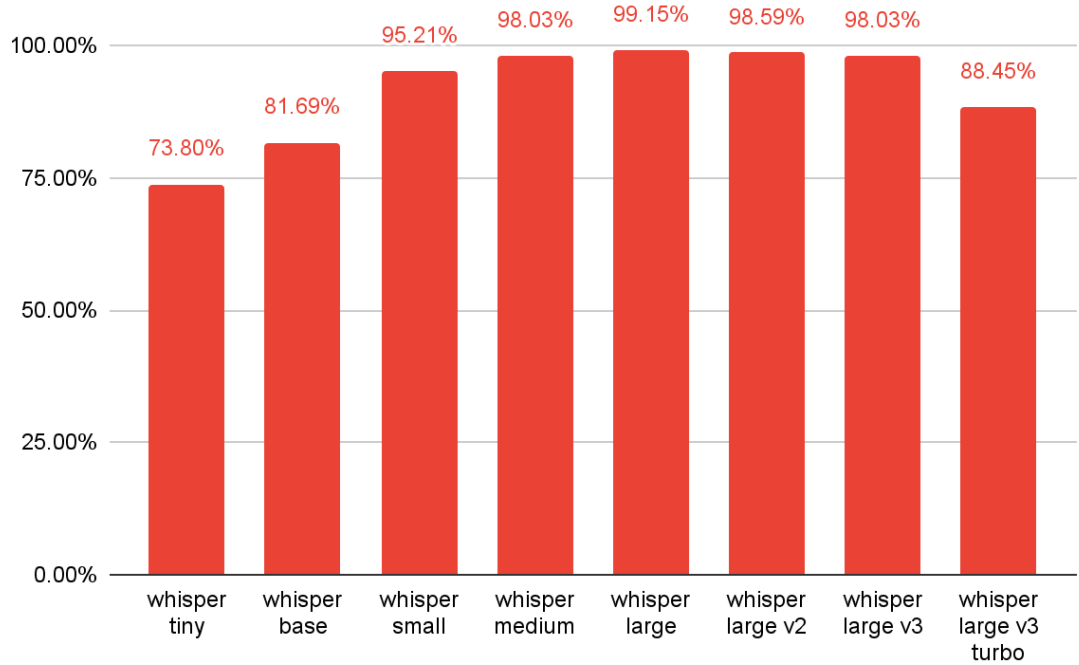

*Figure 4. The language detection accuracy for each of the evaluated Whisper models*

Despite this high language detection accuracy, all models exhibited significant transcription errors on the Arabic Little STT dataset. Detailed CER and WER results are presented in Table 6.

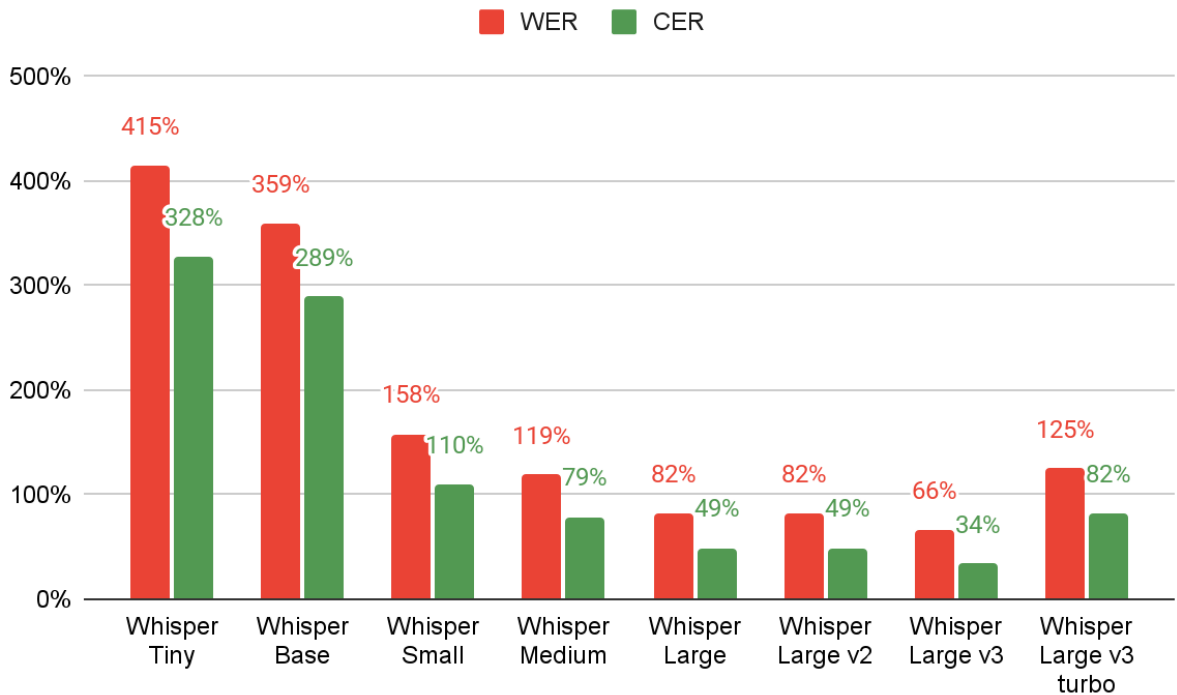

*Figure 5. CER and WER results of the evaluated Whisper models on our dataset*

| Model | WER | CER |
|---|---|---|
| **Whisper Tiny** | 415% | 328% |
| **Whisper Base** | 359% | 289% |
| **Whisper Small** | 158% | 110% |
| **Whisper Medium** | 119% | 79% |
| **Whisper Large** | 82% | 49% |
| **Whisper Large v2** | 82% | 49% |

| Whisper Large v3 | 66% | 34% |
| Whisper Large v3 turbo | 125% | 82% |

*Table 6. WER results of the evaluated Whisper models on our dataset*

A substantial performance drop on child speech can be observed compared to results reported on adult-centric Arabic datasets (Common Voice 9 - Arabic and FLEURS - Arabic) (Radford et al., n.d.). For example, Whisper Large-v3 achieved a WER of 66% on Little STT—4.1× higher than its 16.0% WER on FLEURS (Table 7).

| Model | Our Dataset | Common Voice 9 - Arabic | FLEURS - Arabic |
|---|---|---|---|
| **Whisper Tiny**[11] | 415% | 90.9% | 63.4% |
| **Whisper Base**[12] | 359% | 84.4% | 48.8% |
| **Whisper Small**[13] | 158% | 66.4% | 30.6% |
| **Whisper Medium**[14] | 119% | 60.3% | 20.4% |
| **Whisper Large**[15] | 82% | 56.0% | 18.1% |
| **Whisper Large v2**[16] | **82%** | **53.8%** | **16.0%** |

*Table 7. Comparing the WER result for the evaluated Whisper results on our dataset and the results reported in Whisper Paper (Radford et al., n.d.)*

---

[11] https://huggingface.co/openai/whisper-tiny
[12] https://huggingface.co/openai/whisper-base
[13] https://huggingface.co/openai/whisper-small
[14] https://huggingface.co/openai/whisper-medium
[15] https://huggingface.co/openai/whisper-large
[16] https://huggingface.co/openai/whisper-large-v2

This degradation aligns with Jain et al findings in English child speech evaluations, where Large-v2 performed 20% worse in WER for children compared to adults —Taking the best results for both cases—. The significant increase in transcription errors observed in Arabic, potentially compounded by its linguistic complexity, supports a consistent pattern. Collectively, these results indicate a systemic limitation: ASR models trained primarily on adult data struggle to generalize effectively to children's speech, irrespective of language or model size.

## 5. Conclusion

Arabic children's speech remains a critically underrepresented demographic and linguistic domain in ASR research, hindering advancements in child-centric voice technologies. To overcome this gap, we created a dataset for Arabic children's voices, specifically in Levantine dialect, comprising 355 utterances from 288 children (ages 6–12). We conducted several experiments on our created data by evaluating the performance of Whisper models. Results show a significant performance gap: Whisper's best-performing model (Large v3) achieved a WER of 66% on our dataset, 4.1× higher than its reported WER on adult Arabic benchmarks like FLEUR. This mirrors findings in English child ASR, where Whisper's errors increase by 20% for children compared to adults, suggesting a systemic limitation of current ASR systems. Our work highlights an urgent priority for creating child-inclusive ASR datasets that capture developmental variability. We do hope that our created dataset will help the development of children Arabic ASR. Our future efforts could explore increasing the size of the dataset, or expanding it to cover other dialects to address the challenges in child speech recognition.

## 7. Acknowledgements

We gratefully acknowledge the Genius Planet team for their collaborative spirit and essential support in the collection and curation of the children's speech dataset, a cornerstone of this work.